\documentclass[10pt,twocolumn,letterpaper]{article}

\usepackage{cvpr}      

\usepackage{graphicx}
\usepackage{amsmath}
\usepackage{amssymb}
\usepackage{dsfont}
\usepackage{booktabs}
\usepackage{times}
\usepackage{epsfig}
\usepackage{setspace}
\usepackage{threeparttable}
\usepackage{multirow}
\usepackage{bm}
\usepackage{cite}
\usepackage[ruled]{algorithm}
\usepackage{algpseudocode}
\usepackage{algorithmicx}
\usepackage{adjustbox}
\usepackage[dvipsnames,table,xcdraw]{xcolor}
\usepackage{pifont}
\usepackage[accsupp]{axessibility}
\usepackage[pagebackref=false,breaklinks,colorlinks,citecolor=RoyalBlue,bookmarks=false]{hyperref}

\newcommand{\highlight}[1]{\textbf{\textcolor{ForestGreen}{#1}}}

\newlength\savewidth\newcommand\shline{\noalign{\global\savewidth\arrayrulewidth
  \global\arrayrulewidth 1pt}\hline\noalign{\global\arrayrulewidth\savewidth}}
\newcommand{\tablestyle}[2]{\setlength{\tabcolsep}{#1}\renewcommand{\arraystretch}{#2}\centering\footnotesize}
\makeatletter\renewcommand\paragraph{\@startsection{paragraph}{4}{\z@}
  {.5em \@plus1ex \@minus.2ex}{-.5em}{\normalfont\normalsize\bfseries}}\makeatother
  
\newcommand{\myPara}[1]{\vspace{.05in}\noindent\textbf{#1}}
  
\usepackage[capitalize]{cleveref}
\crefname{section}{Sec.}{Secs.}
\Crefname{section}{Section}{Sections}
\Crefname{table}{Table}{Tables}
\crefname{table}{Tab.}{Tabs.}

\def\cvprPaperID{1767} 
\def\confName{CVPR}
\def\confYear{2022}

\begin{document}

\title{Localization Distillation for Dense Object Detection}

\author{{Zhaohui Zheng$^{1*}$,~ Rongguang Ye$^{2}$\thanks{Equal contribution.}} ,~ Ping Wang$^{2}$,~ Dongwei Ren$^{3}$, ~ Wangmeng Zuo$^3$,\\ ~ Qibin Hou$^{1}$\thanks{Corresponding author.} , ~ Ming-Ming Cheng$^1$ \\
  \textsuperscript{\rm 1}TMCC, CS, Nankai University \qquad
		\textsuperscript{\rm 2}School of Mathematics, Tianjin University\\
		\textsuperscript{\rm 3}School of Computer Science and Technology, Harbin Institute of Technology \\
	}
\maketitle

\begin{abstract} 
Knowledge distillation (KD) has witnessed its powerful capability 
in learning compact models in object detection.
Previous KD methods for object detection mostly focus on imitating deep features
within the imitation regions instead of mimicking classification logit
due to its inefficiency in distilling localization information and trivial improvement.
In this paper, by reformulating the knowledge distillation process on localization,
we present a novel localization distillation (LD) method which can efficiently
transfer the localization knowledge from the teacher to the student.
Moreover, we also heuristically introduce the concept of valuable localization region
that can aid to selectively distill the semantic and localization knowledge
for a certain region.
Combining these two new components, for the first time, we show that 
logit mimicking can outperform feature imitation and 
localization knowledge distillation is more important and efficient than 
semantic knowledge for distilling object detectors.
Our distillation scheme is simple as well as effective and can be easily applied to different dense object detectors.
Experiments show that our LD can boost the AP score of GFocal-ResNet-50
with a single-scale 1$\times$ training schedule from 40.1 to 42.1 
on the COCO benchmark without any sacrifice on the inference speed.
Our source code and pretrained models are publicly available at \url{https://github.com/HikariTJU/LD}.

\end{abstract}

\section{Introduction}

Localization is a fundamental issue in object detection \cite{locnet,wang2019region,SABL,zhu2019feature,gridrcnn,kong2018deep,SCRDet,GWD,KLD,VFNet}.
Bounding box regression is the most popular manner so far for localization 
in object detection \cite{felzenszwalb2009object,yolov1,SSD,fasterrcnn}, 
where Dirac delta distribution representation is intuitive and popular for years.
\begin{figure}[!t]
	\centering
	\setlength{\tabcolsep}{1pt}
	\setlength{\abovecaptionskip}{3pt}
	\begin{tabular}{cccccccccc}
		\includegraphics[width=0.2\textwidth]{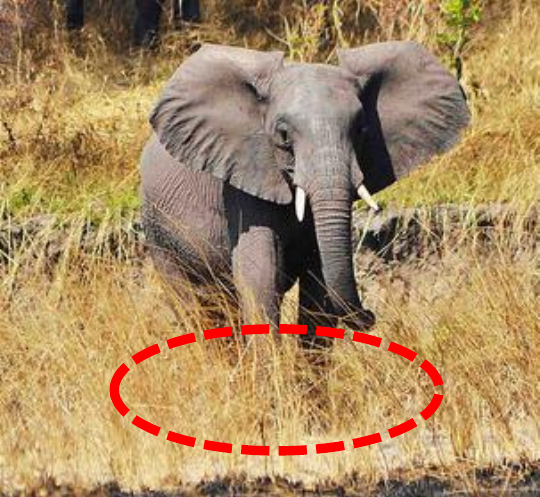}&
		\includegraphics[width=0.22\textwidth]{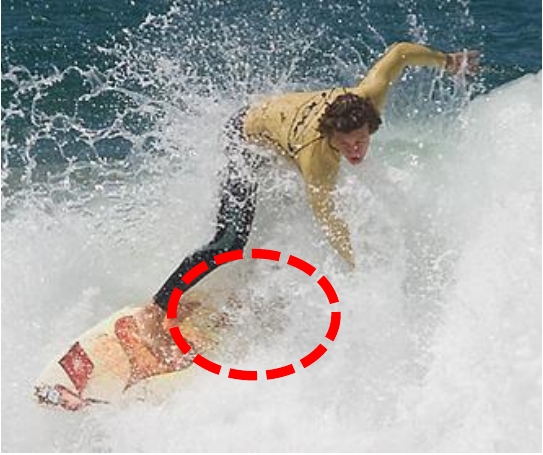}\\
	\end{tabular}
	\caption{Bottom edge for ``\emph{elephant}" and right edge for ``\emph{surfboard}" are ambiguous.
		}
	\label{fig:ambiguity}
\end{figure}
However, localization ambiguity where objects cannot be 
confidently located by their edges is still a common issue.
For example, as shown in Fig. \ref{fig:ambiguity}, the bottom edge 
for ``elephant" and the right edge for ``surfboard" are ambiguous to locate.
This issue is even worse for lightweight detectors.
One way to alleviate this problem is the knowledge distillation (KD), which, as a model compression technology, has been widely validated to be useful for boosting the performance of the small-sized student network by transferring the generalized knowledge captured by the large-sized teacher network.
	
\begin{figure}[!t]
	\centering
	\setlength{\tabcolsep}{1pt}
	\setlength{\abovecaptionskip}{3pt}
	\begin{tabular}{cccccccccc}
		\includegraphics[width=0.46\textwidth]{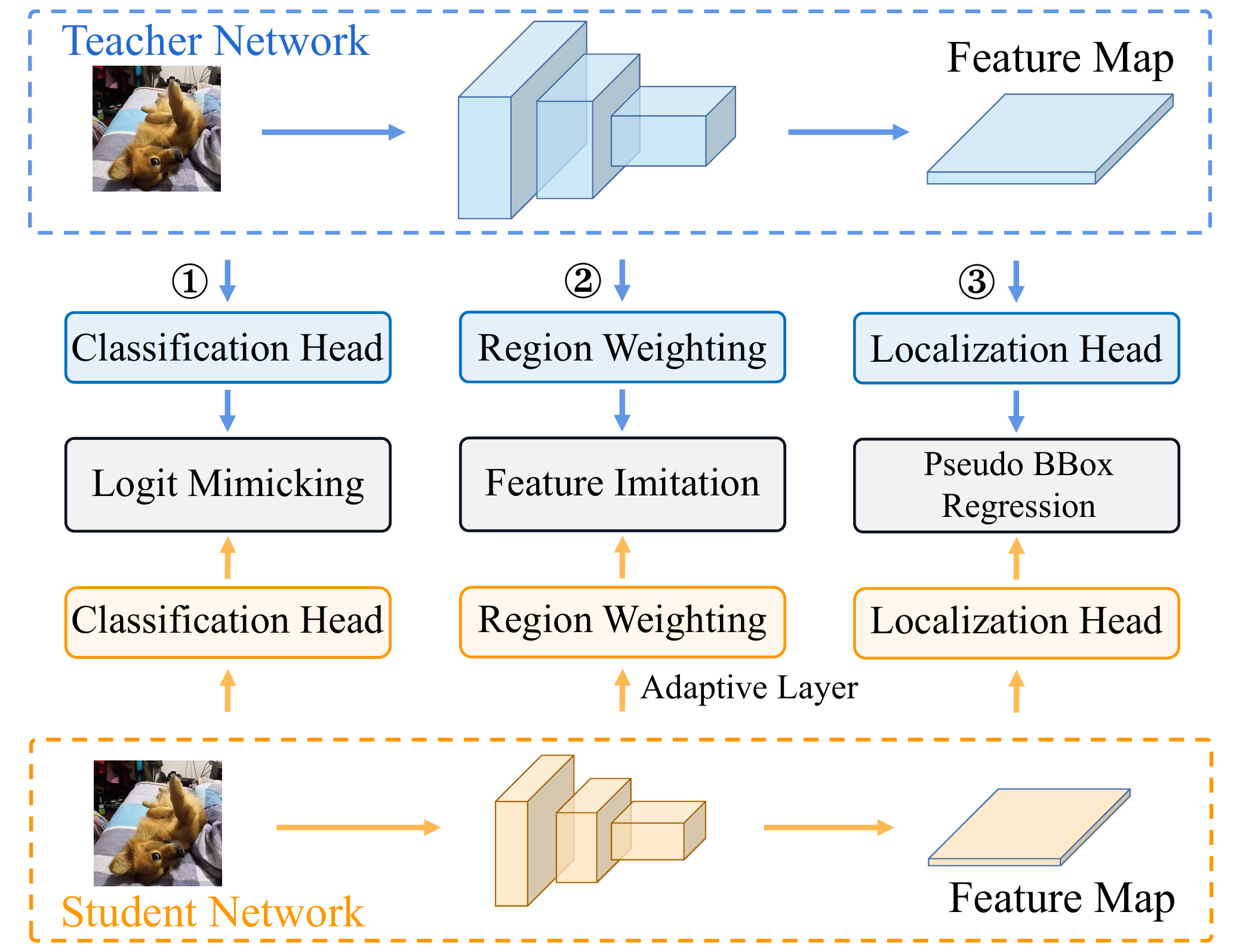}\\
	\end{tabular}
	\caption{Existing KD pipelines for object detection. \ding{172} Logit Mimicking: classification KD in \cite{hinton2015distilling}. \ding{173} Feature Imitation: recent popular methods distill intermediate features based on various distillation regions, which usually need adaptive layers to align the size of the student's feature map. \ding{174} Pseudo BBox Regression: treating teachers' predicted bounding boxes as additional regression targets.
		}
	\label{fig:previous}
\end{figure}

Speaking of KD in object detection, previous works \cite{wang2019distilling,zhang2020improve,kang2021instanceconditional} 
have pointed out that the original logit mimicking technique \cite{hinton2015distilling}
for classification is inefficient as it only transfers the semantic knowledge (\ie, classification),
while neglects the importance of localization knowledge distillation.
Therefore, existent KD methods for object detection mostly focus on enforcing 
the consistency of the deep features between the teacher-student pair,
and exploit various imitation regions for distillation \cite{chen2017learning,Li_2017_CVPR,wang2019distilling,GIbox,defeat}.
Fig.~\ref{fig:previous} exhibits three popular KD pipelines for object detection.
However, as the semantic knowledge and the localization one are mixed on the feature maps,
it is hard to tell whether it is beneficial to the performance 
to transfer the hybrid knowledge for each location and
which regions are conducive to the transfer of a certain type of knowledge.

Motivated by the aforementioned questions, in this paper, instead of simply distilling the hybrid knowledge on the feature maps, we propose a novel divide-and-conquer distillation strategy that transfers the semantic and localization knowledge separately.
For semantic knowledge, we use the original classification KD \cite{hinton2015distilling}.
For localization knowledge, we reformulate the knowledge transfer process on localization
and present a simple yet effective localization distillation (LD) method by switching the bounding box to probability distribution \cite{offsetbin,gfocal}.
This is quite different from previous works \cite{chen2017learning,sun2020distilling} 
that treat the teacher's outputs as additional regression 
targets (\ie, the Pseudo BBox Regression in Fig.~\ref{fig:previous}).
Benefiting from the probability distribution representation, our LD can efficiently
transfer rich localization knowledge learnt by the teacher to the student.
In addition, based on the proposed divide-and-conquer distillation strategy, 
we further introduce valuable localization region (VLR)
to help efficiently judge which regions are conducive to classification 
or localization learning.
Through a series of experiments, \emph{we, for the first time, show that the original logit mimicking 
can be better than feature imitation and localization knowledge distillation 
is more important and more efficient than semantic knowledge.}
We believe that separately distilling the semantic and localization knowledge 
based on their respective favorable regions could be a promising way to 
train better object detectors.

Our method is simple and can be easily equipped with in any 
dense object detectors to improve their performance without introducing any inference overhead.
Extensive experiments on MS COCO show that
without bells and whistles, we can lift the AP score of 
the strong baseline GFocal~\cite{gfocal} with ResNet-50-FPN backbone from 40.1 to 42.1, and AP$_{75}$ from 43.1 to 45.6.
Our best model using ResNeXt-101-32x4d-DCN backbone can achieve a single-scale test of 50.5 AP, which surpasses all existing detectors under the same backbone, neck, and test settings.

\section{Related Work}\label{sec:related}
	
In this section, we give a brief review on the related works,
including bounding box regression, localization quality estimation,
and knowledge distillation.

\subsection{Bounding Box Regression}
Bounding box regression is the most popular method 
for localization in object detection.
R-CNN series \cite{fasterrcnn,cascadercnn,librarcnn,DynamicRCNN} 
adopt multiple regression stages to refine the detection results, 
while \cite{yolov1,yolov2,yolov3,yolov4,SSD,FCOS} adopt 
one-stage regression.
In \cite{unitbox,giou,diou,ciou}, IoU-based loss functions are proposed 
to improve the localization quality of bounding box.
Recently, bounding box representation has evolved from Dirac delta distribution \cite{yolov1,SSD,fasterrcnn} to Gaussian distribution \cite{softernms,gaussian_yolov3}, and further to probability
distribution \cite{offsetbin,gfocal}.
The probability distribution of bounding box is more comprehensive 
for describing the uncertainty of bounding box, and is validated to 
be the most advanced bounding box representation so far.

\subsection{Localization Quality Estimation}
As the name suggests, Localization Quality Estimation (LQE) predicts a score that measures the localization quality of the bounding box predicted by the detector.
LQE is usually used to cooperate with classification task during training \cite{gfocalv2}, \ie, enhancing the consistency between classification and localization.
It can also be applied in joint decision-making during post-processing \cite{yolov1,iounet,FCOS}, \ie, considering both classification score and LQE when performing NMS.
Early research can be dated to YOLOv1 \cite{yolov1}, where the predicted object confidence is used to penalize the classification score.
Then, box/mask IoU \cite{iounet,mask_scoring} and box/polar center-ness \cite{FCOS,polarmask} are proposed to model the uncertainty of detections for object detection and instance segmentation, respectively.
From the perspective of bounding box representation, Softer-NMS \cite{softernms} and Gaussian YOLOv3 \cite{gaussian_yolov3} predict variance for each edge of bounding box.
LQE is a preliminary approach to model localization ambiguity.

\begin{figure*}[!t]
    \centering
    \setlength{\abovecaptionskip}{5pt}
    \includegraphics[width=1\textwidth]{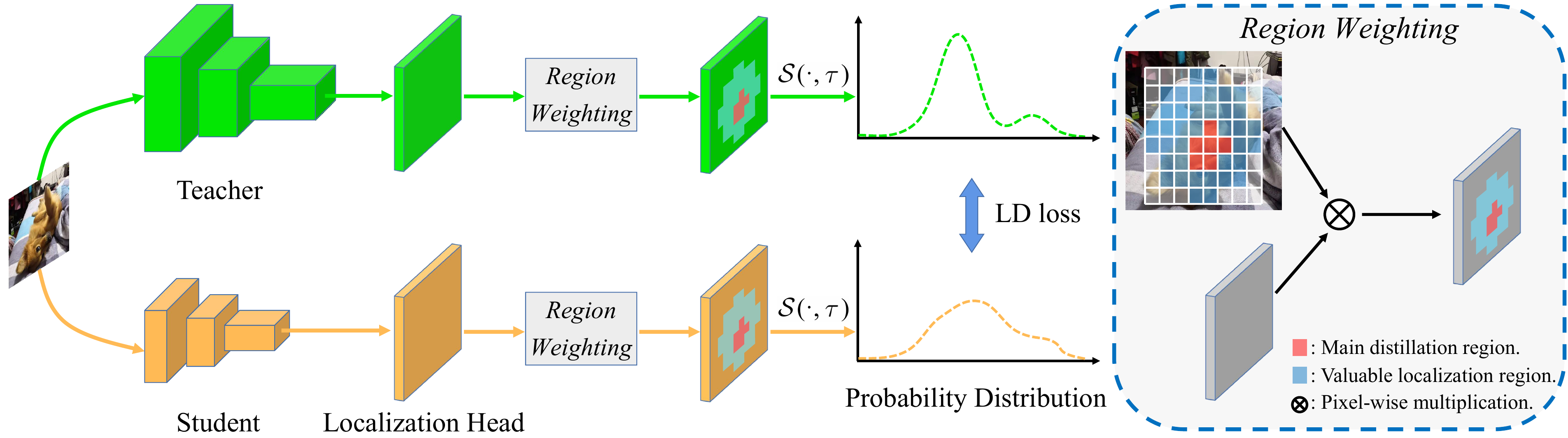}
    \caption{Illustration of localization distillation (LD) for an edge $e \in \mathcal{B} =\{t,b,l,r\}$. Only the localization branch is visualized here. $\mathcal{S}(\cdot,\tau)$ is the generalized SoftMax function with temperature $\tau$. For a given detector, we first switch the bounding box representation to probability distribution. Then, we determine where to distill via region weighting on the main distillation region and the valuable localization region. Finally, we calculate the LD loss between two probability distributions predicted by the teacher and the student.}
    \label{fig:LD}
    \end{figure*}	
\subsection{Knowledge Distillation}

Knowledge distillation \cite{hinton2015distilling,Zagoruyko2017AT,bae2020densely,relationKD,TA,DenselyTA} aims to learn compact and efficient student models guided by excellent teacher networks.
FitNets \cite{FitNets} proposes to mimic the intermediate-level hints
from the hidden layers of the teacher model.
Knowledge distillation was first applied to object detection in \cite{chen2017learning}, where the hint learning and KD are both 
used for multi-class object detection.
Then,  Li et al. \cite{Li_2017_CVPR} proposed to mimic the feature within the region proposal for Faster R-CNN.
Wang et al.~\cite{wang2019distilling} mimicked fine-grained features
on close anchor box locations.
Recently, Dai et al.~\cite{GIbox} introduced the General Instance Selection Module to mimic deep features within the discriminative patches between teacher-student pairs.
DeFeat \cite{defeat} leverages different loss weights when conducting feature imitation on the object regions and the background region.
Different from the aforementioned method based on feature imitation, 
our work introduces localization distillation and proposes to separately
transfer the classification and localization knowledge based on 
the valuable localization region to make the distillation more efficient.

\section{Proposed Method}
In this section, we introduce the proposed distillation method.
Instead of distilling the hybrid knowledge on feature maps, 
we present a novel divide-and-conquer distillation strategy that
separately distills the semantic and localization knowledge 
based on their respective preferred regions.
To transfer the semantic knowledge, we simply adopt the classification 
KD \cite{hinton2015distilling} on the classification head 
while for localization knowledge, we propose a simple yet effective 
localization distillation (LD).
Both techniques operate on the logits of individual heads rather than deep features.
Then, to further improve the distillation efficiency, 
we introduce valuable localization region (VLR) that can help judge 
which type of knowledge is conducive to transfer for different regions.
In what follows, we first briefly revisit the probability distribution representation of bounding box and then transit to the proposed method.

\subsection{Preliminaries}

For a given bounding box $\mathcal{B}$, conventional representations
have two forms, \ie, $\{x,y,w,h\}$ (central point coordinates, width and height)~\cite{yolov1,SSD,fasterrcnn} and $\{t,b,l,r\}$ (distance from the sampling point to the top, bottom, left and right edges)~\cite{FCOS}.
These two forms actually follow the Dirac delta distribution that only focuses on the ground-truth locations but cannot model the ambiguity of bounding boxes 
as shown in Fig.~\ref{fig:ambiguity}.
This is also clearly demonstrated in  previous works ~\cite{softernms,gaussian_yolov3,offsetbin,gfocal}.

In our method, we use the recent probability distribution representation of bounding box \cite{offsetbin,gfocal} which is more comprehensive for describing the localization uncertainty of bounding box.
Let $e \in \mathcal{B}$ be an edge of a bounding box. Its value can be generally represented as
\begin{equation}\label{eq:general}
	\hat{e}=\int_{e_{\min}}^{e_{\max}} x \Pr(x) dx,\quad e\in \mathcal{B},
\end{equation}
where $x$ is the regression coordinate ranged in $[e_{\min}, e_{\max}]$, and $\Pr(x)$ is the corresponding probability.
The conventional Dirac delta representation is a special case of Eqn. \eqref{eq:general}, where $\Pr(x) =1 \text{ when } x = e^{gt}$, otherwise $\Pr(x) = 0$.
By quantizing the continuous regression range $[e_{\min}, e_{\max}]$ into the uniform discretized variable $\bm{e}=[e_1,e_2,\cdots,e_n]^\text{T} \in \mathbb{R}^n$ with $n$ subintervals, where $e_1=e_{\min}$ and $e_n=e_{\max}$, each edge of the given bounding box can be represented as the probability distribution
by using the SoftMax function.

\subsection{Localization Distillation} \label{sec:LD}

In this subsection, we present localization distillation (LD), a new way to
enhance the distillation efficiency for object detection.
Our LD is evolved from the view of probability distribution representation
~\cite{gfocal} of bounding box which is originally designed for the generic
object detection and carries abundant localization information.
The ambiguous and clear edges in Fig. \ref{fig:ambiguity} will 
be respectively reflected by the flatness and sharpness of distribution.

The working principle of our LD can be seen in Fig. \ref{fig:LD}.
Given an arbitrary dense object detector, following \cite{gfocal}, we first switch the bounding box representation from a quaternary representation to a probability distribution.
We choose $\mathcal{B} = \{t,b,l,r\}$ as the basic form for bounding box.
Unlike the $\{x,y,w,h\}$ form, the physical meaning of each variable 
in the $\{t,b,l,r\}$ form is consistent, which is convenient for us to 
restrict the probability distribution of each edge to the same interval range.
According to \cite{ATSS}, there is no performance difference between the two forms.
Thus, when the $\{x,y,w,h\}$ form is given, we will first switch it 
to the $\{t,b,l,r\}$ form.

Let $\bm{z}$ be the $n$ logits predicted by the localization head 
for all possible positions of edge $e$, denoted by $\bm{z}_T$ and $\bm{z}_S$
for the teacher and the student, respectively.
Different from \cite{offsetbin,gfocal}, we transform $\bm{z}_T$ and $\bm{z}_S$ into probability distributions $\bm{p}_T$ and $\bm{p}_S$ using the generalized SoftMax function $\mathcal{S}(\cdot,\tau)=\text{SoftMax}(\cdot/\tau)$.
Note that when $\tau=1$, it is equivalent to the original SoftMax function.
When $\tau\rightarrow0$, it tends to be a Dirac delta distribution.
When $\tau\rightarrow\infty$, it will degrade to be a uniform distribution.
Empirically, $\tau > 1$ is set to soften the distribution, making the probability distribution carry more information.

The localization distillation for measuring the similarity 
between the two probabilities $\bm{p}_T, \bm{p}_S\in\mathbb{R}^n$ is attained by:
\begin{align}
	\mathcal{L}_\text{LD}^e &= \mathcal{L}_\text{KL}(\bm{p}_S^\tau,\bm{p}_T^\tau) \\
	                        &= \mathcal{L}_\text{KL}\left(\mathcal{S}(\bm{z}_S, \tau), \mathcal{S}(\bm{z}_T, \tau)\right),
\end{align}
where $\mathcal{L}_\text{KL}$ represents the KL-Divergence loss.
Then, LD for all the four edges of bounding box $\mathcal{B}$ can be formulated as:
\begin{equation}
	\mathcal{L}_\text{LD} (\mathcal{B}_S, \mathcal{B}_T)= \sum_{e \in \mathcal{B}} \mathcal{L}_\text{LD}^{e}.
\end{equation}

\noindent\textbf{Discussion.}
Our LD is the first attempt to adopt logit mimicking to distill localization
knowledge for object detection.
Though the probability distribution representation for boxes has been proven
useful in the generic object detection task~\cite{gfocal}, no one has explored
its performance in localization knowledge distillation.
We combine the probability distribution representation for boxes and
the KL-Divergence loss and demonstrate that such a simple logit mimicking technique
performs well in improving the distillation efficiency of object detectors.
This also makes our LD quite different from previous relevant works that, 
on the contrary, emphasize the importance of feature imitation.
In our experiment section,we will show more numerical analysis on the advantages
of the proposed LD.
\subsection{Valuable Localization Region}
\begin{algorithm}[!tb]	
	\footnotesize
	\caption{Valuable Localization Region}
	\label{alg:VLR}
	\begin{algorithmic}[1]
		\small{	
			\Require{A set of anchor boxes $\bm{B}_l^a = \{\mathcal{B}^a_{i_l}\}$ and a set of ground truth boxes $\bm{B}^{gt} = \{\mathcal{B}^a_{j}\}$, $1\leqslant i_l\leqslant I_l$ , $1\leqslant j\leqslant J$, $I_l=W_l\times H_l$. Positive threshold $\alpha_{pos}$ of label assignment. $W_l$ and $H_l$ are the sizes of $l$-th FPN level.
			}
			\Ensure{$\bm{V}_l=\{v_{i_lj}\}_{I_l\times J}, v_{i_lj}\in \{0,1\}$ encodes final location of VLR, where $1$ denotes VLR and $0$ indicates ignore.}}
		\State Compute DIoU matrix $\bm{X}_l=\{x_{i_lj}\}_{I_l\times J}$ with $x_{i_lj}=DIoU(\mathcal{B}^a_{i_l},\mathcal{B}^{gt}_j)$.
		\State $\alpha_{vl}=\gamma\alpha_{pos}$.
		\State Select locations with $\bm{V}_l=\{\alpha_{vl}\leqslant\bm{X}_l\leqslant\alpha_{pos}\}$.
		\State\textbf{return} {$\bm{V}_l$}
	\end{algorithmic}
\end{algorithm}

\begin{table*}[t]
\centering
\subfloat[\footnotesize \textbf{Temperature $\tau$ in LD}: The generalized Softmax function with large $\tau$ brings considerable gains. We set $\tau$$=$$10$ by default. The teacher is ResNet-101 and the student is ResNet-50.\label{tab3a}]{
\tablestyle{3pt}{1.2}
\footnotesize 
\begin{tabular}{c|c c c|c c c}
$\tau$ & AP & AP$_{50}$ & AP$_{75}$ & AP$_{S}$ & AP$_{M}$ & AP$_{L}$  \\
\shline
 -- & 40.1 & 58.2 & 43.1 & 23.3 & 44.4 & 52.5 \\
  \hline
 1 & 40.3 & 58.2 & 43.4 & 22.4 & 44.0 & 52.4 \\
 5  & 40.9 & 58.2 & 44.3 & 23.2 & \highlight{45.0} & 53.2\\
10 & \highlight{41.1} & \highlight{58.7} & \highlight{44.9} & \highlight{23.8} & 44.9 & \highlight{53.6}\\
 15 & 40.7 & 58.5 & 44.2 & 23.5 & 44.3 & 53.3 \\
 20 & 40.5 & 58.3 & 43.7 & \highlight{23.8} & 44.1 & 53.5
\end{tabular}} \hspace{3mm}
\subfloat[\footnotesize  \textbf{LD {\em vs}. Pseudo BBox Regression \cite{chen2017learning}}: The localization knowledge can be more efficiently transferred by our LD. The teacher is ResNet-101 and the student is ResNet-50.\label{tab3b}]{
\tablestyle{3pt}{1.2}
\footnotesize 
\begin{tabular}{c|c c c|c c c}
$\varepsilon$ & AP & AP$_{50}$ & AP$_{75}$ & AP$_{S}$ & AP$_{M}$ & AP$_{L}$  \\
 \shline
-- & 40.1 & 58.2 & 43.1 & 23.3 & 44.4 & 52.5 \\
 \hline
 0.1 & 40.5  & 58.3 & 43.8 &  23.0 &  44.2 &  52.7\\
 0.2 & 40.2 & 58.2 & 43.6 &  23.1 &  44.0 & 53.0 \\
 0.3 & 40.1 & 58.4 & 43.1 & 23.6 & 43.9 & 52.5 \\
 0.4  & 40.3 & 58.4 & 43.4 & 22.8 & 44.0 & 52.6 \\
 \hline
 LD  &  \highlight{41.1} & \highlight{58.7} & \highlight{44.9} & \highlight{23.8} & \highlight{44.9} & \highlight{53.6}\\
\end{tabular}}\hspace{3mm}
\subfloat[\footnotesize \textbf{Role of $\gamma$ in VLR}: Conducting LD on valuable localization region has a positive effect on performance. We set $\gamma$$=$$0.25$ by default.
The teacher is ResNet-101 and the student is ResNet-50.\label{tab3c}]{
\tablestyle{3pt}{1.2}
\footnotesize
\begin{tabular}{c|c c c|c c c}
$\gamma$ & AP & AP$_{50}$ & AP$_{75}$ & AP$_{S}$ & AP$_{M}$ & AP$_{L}$  \\
 \shline
 -- & 40.1 & 58.2 & 43.1 & 23.3 & 44.4 & 52.5 \\
 \hline
 1 & 41.1 & 58.7 & 44.9 & 23.8 & 44.9 & 53.6\\
 0.75  & 41.2 & 58.8 & 44.9 & 23.6 & 45.4 & 53.5\\
 0.5 & 41.7 & 59.4 & 45.3 & 24.2 & 45.6 & 54.2\\
 0.25 & \highlight{41.8} & \highlight{59.5} & \highlight{45.4} & 24.2 & 45.8 & \highlight{54.9} \\
 0 & 41.7 & \highlight{59.5} & \highlight{45.4} & \highlight{24.5} & \highlight{45.9} & 54.0
\end{tabular}}\hspace{3mm}
\caption{\textbf{Ablations}. We show ablation experiments for LD and VLR on MS COCO val2017.}
\label{tab3}
\end{table*}
Previous works mostly force the deep features of 
the student to mimic those of the teacher 
by minimizing the $l_2$ loss.
However, a straightforward question should be: Should we use 
the whole imitation regions without discrimination to distill the hybrid knowledge?
According to our observation, the answer is no.
Previous works \cite{song2020revisiting,HTD,gao2021mutual,TOOD,wang2021reconcile} 
have pointed out that the knowledge distribution patterns are different for classification and localization.
Therefore, we, in this subsection, describe the valuable localization region (VLR), to further improve the distillation efficiency,
which we believe will be a promising way to train better student detectors.

Specifically, the distillation region is divided into two parts, the main distillation region and the valuable localization region.
The main distillation region is intuitively determined by label assignment, \ie, the positive locations of the detection head.
The valuable localization region can be obtained by Algorithm \ref{alg:VLR}.
First, for the $l$-th FPN level, we calculate the DIoU \cite{diou} matrix $\bm{X}_l$ between all the anchor boxes $\bm{B}_l^a$ and the ground-truth boxes $\bm{B}^{gt}$.
Then, we set the lower bound of DIoU to be $\alpha_{vl}=\gamma\alpha_{pos}$, where $\alpha_{pos}$ is the positive IoU threshold of label assignment.
The VLR can be defined as $\bm{V}_l=\{\alpha_{vl}\leqslant\bm{X_l}\leqslant\alpha_{pos}\}$.
Our method has only one hyperparameter $\gamma$, which controls the range of the VLRs.
When $\gamma=0$, all the locations whose DIoUs  
between the preset anchor boxes and the GT boxes satisfy $0\leqslant x_{i_lj}\leqslant\alpha_{pos}$ will be determined as VLRs.
When $\gamma\rightarrow1$, the VLR will gradually shrink to empty.
Here we use DIoU \cite{diou} since it gives higher priority to the locations close to the center of the object.

Similar to label assignment, our method assigns attributes to each location across multi-level FPN.
In this way, some of locations outside GT boxes will also be considered.
So, we can actually view the VLR as 
an outward extension of the main distillation region.
Note that for anchor-free detectors, like FCOS, we can use the preset anchors on feature maps and do not change its regression form so that the localization learning maintains to be anchor-free type.
While for anchor-based detectors which usually set multiple anchors per location, we unfold the anchor boxes to calculate the DIoU matrix, and then assign their attributes.

\subsection{Overall Distillation Process}

The total loss for training the student $\bm S$ can be represented as:
\begin{equation} \label{eqn:total_loss}
\begin{aligned}
	\mathcal{L}=&\lambda_0\mathcal{L}_\text{cls}(\mathcal{C}_S,\mathcal{C}^{gt})+\lambda_1\mathcal{L}_\text{reg}(\mathcal{B}_S,\mathcal{B}^{gt})+\lambda_2\mathcal{L}_\text{DFL}(\mathcal{B}_S,\mathcal{B}^{gt})\\
                          +&\lambda_3\mathds{I}_\text{Main}\mathcal{L}_\text{LD}(\mathcal{B}_S, \mathcal{B}_T)+\lambda_4\mathds{I}_\text{VL}\mathcal{L}_\text{LD}(\mathcal{B}_S, \mathcal{B}_T)\\
                          +&\lambda_5\mathds{I}_\text{Main}\mathcal{L}_\text{KD}(\mathcal{C}_S, \mathcal{C}_T)+\lambda_6\mathds{I}_\text{VL}\mathcal{L}_\text{KD}(\mathcal{C}_S, \mathcal{C}_T),
\end{aligned}
\end{equation}
where the first three terms are exactly same to the classification and bounding box regression branches for any regression-based detector, \ie, $\mathcal{L}_\text{cls}$ is the classification loss, $\mathcal{L}_\text{reg}$ is the bounding box regression loss and $\mathcal{L}_\text{DFL}$ is the distribution focal loss \cite{gfocal}. $\mathds{I}_\text{Main}$ and $\mathds{I}_\text{VL}$ are the distillation masks for the main distillation region and the valuable localization region respectively, $\mathcal{L}_\text{KD}$ is KD loss~\cite{hinton2015distilling}, $\mathcal{C}_S$ and $\mathcal{C}_T$ denote the classification head output logits of the student and the teacher, respectively, $\mathcal{C}^{gt}$ is the ground truth class label.
All the distillation losses will be weighted by the same weight factors according to their types, \eg, LD loss follows the bbox regression and KD loss follows the classification.
Also, it is worth mentioning that DFL loss term can be disabled since LD loss has sufficient guidance ability.
\emph{In addition, we can enable or disable the four types of distillation losses so as to distill the student in a separate distillation region manner.}
	
\section{Experiment} \label{sec:results}

In this section, we conduct comprehensive ablation studies and analysis to demonstrate the superiority of the proposed LD and distillation scheme on the challenging large-scale MS COCO \cite{coco} benchmark.

\subsection{Experiment Setup}

The train2017 (118K images) is utilized for training and val2017 (5K images) is used for validation.
We also obtain the evaluation results on MS COCO test-dev 2019 (20K images) by submitting to the COCO server.
The experiments are conducted under mmDetection \cite{mmdetection} framework.
Unless otherwise stated, we use ResNet \cite{ResNet} with FPN \cite{FPN} as our backbone and neck networks, and the FCOS-style \cite{FCOS} anchor-free head for classification and localization.
The training schedule for ablation experiments is set to single-scale 1$\times$ mode (12 epochs).
For other training and testing hyper-parameters, we follow exactly the GFocal \cite{gfocal} protocol, including QFL loss for classification and GIoU loss for bbox regression \etc.
We use the standard COCO-style measurement, \ie, average precision (AP), for evaluation.
All the baseline models are retrained by adopting the same settings so as to fairly compare them with our LD.
More implementation details and more experimental results on PASCAL VOC \cite{voc} can be found in the supplementary materials.

\subsection{Ablation Studies and Analysis}

\myPara{Temperature $\tau$ in LD.}
Our LD introduces a hyper-parameter, \ie, the temperature $\tau$.
Table \ref{tab3a} reports the results of LD with various temperatures, 
where the teacher model is ResNet-101 with AP 44.7 and the student model is ResNet-50.
Here, only the main distillation region is adopted.
Compared to the first row in Table \ref{tab3a}, different temperatures
consistently lead to better results.
In this paper, we simply set the temperature in LD as $\tau=10$, which is fixed in all the other experiments.

\myPara{LD {\em vs}. Pseudo BBox Regression.}
The teacher bounded regression (TBR) loss~\cite{chen2017learning} is a preliminary attempt to enhance the student on the localization head, \ie, the pseudo bbox regression in Fig.~\ref{fig:previous},
which is represented as:
\begin{equation} \label{eq:tbr}
 	\begin{aligned}
 		\mathcal{L}_\text{TBR}=\lambda\mathcal{L}_\text{reg}(\mathcal{B}^{s},\mathcal{B}^{gt}),\text{if}\;\ell_2(\mathcal{B}^{s},\mathcal{B}^{gt})+\varepsilon > \ell_2(\mathcal{B}^{t},\mathcal{B}^{gt}),\\
 	\end{aligned}
\end{equation}
where $\mathcal{B}^{s}$ and $\mathcal{B}^{t}$ denote the predicted boxes of student and teacher respectively, $\mathcal{B}^{gt}$ denotes the ground truth boxes, $\varepsilon$ is a predefined margin, and $\mathcal{L}_\text{reg}$ represents the GIoU loss \cite{giou}.
Here, only the main distillation region is adopted.
From Table \ref{tab3b}, TBR loss does yield performance gains (+0.4 AP and +0.7 AP$_{75}$) when using proper threshold $\varepsilon=0.1$ in Eqn. \eqref{eq:tbr}.
However, it uses the coarse bbox representation, which does not contain any localization uncertainty information of the detector, leading to sub-optimal results.
On the contrary, our LD directly produces 41.1 AP and 44.9 AP$_{75}$, since it utilizes the probability distribution of bbox which contains rich localization knowledge.

\myPara{Various $\gamma$ in VLR.}
The newly introduced VLR has the parameter $\gamma$ which controls the range of VLR.
As shown in Table \ref{tab3c}, AP is stable when $\gamma$ ranges from 0 to 0.5.
The variation in AP in this range is around 0.1.
As $\gamma$ increases, the VLR gradually shrinks to empty.
The performance also gradually drops to 41.1, \ie, conducting LD on the main distillation region only.
The sensitivity analysis experiments on the parameter $\gamma$ indicate
that conducting LD on the VLR has a positive effect on performance.
In the rest experiments, we set $\gamma$ to 0.25 for simplicity.

\begin{table}[!t]\small
	\centering
	\setlength{\abovecaptionskip}{2pt}
	\renewcommand\arraystretch{1.1}
	\caption{Evaluation of \textbf{separate distillation region manner} for KD and our LD. The teacher is ResNet-101 and the student is ResNet-50. ``Main" denotes the main distillation region, \ie, the positive locations of label assignment. ``VLR" denotes the valuable localization region. The results are reported on MS COCO val2017.}
	\begin{spacing}{1.1}
		\resizebox{0.47\textwidth}{!}{
			\begin{tabular}{cccc|c c c}
				\hline
				
				\hline
				 Main KD & Main LD & VLR KD & VLR LD & AP & AP$_{50}$ & AP$_{75}$   \\
                \hline
				 &&&& 40.1 & 58.2 & 43.1  \\
                \hline
                 \checkmark &&&& 40.2 & 58.6 & 43.4  \\
                 & \checkmark &&& 41.1 & 58.7 & 44.9 \\
                \checkmark & \checkmark &&& 41.4 & 59.2 & 45.0  \\
                 \checkmark && \checkmark && 40.4 & 58.9 & 43.4  \\
                 & \checkmark && \checkmark & 41.8 & 59.5 & 45.4  \\
                \checkmark & \checkmark && \checkmark  & \highlight{42.1} & \highlight{60.3} & \highlight{45.6} \\
                \checkmark & \checkmark & \checkmark & \checkmark  & 42.0 & 60.0 & 45.4  \\
				\hline
				
				\hline
			\end{tabular}}
	\end{spacing}

	\label{tab:separateKD}
\end{table}

\myPara{Separate Distillation Region Manner.}
There are several interesting observations regarding the roles of KD and LD and their preferred regions.
We report the relevant ablation study results in Table \ref{tab:separateKD}, 
where ``Main" indicates that the logit mimicking is conducted on the main distillation region, \ie, the positive locations of label assignment, and ``VLR” denotes the valuable localization region.
It can be seen that conducting ``Main KD", ``Main LD", and their combination
can all improve the student performance by +0.1, +1.0 and +1.3 AP, respectively.
This indicates that the main distillation regions contain the valuable knowledge
for both classification and localization and the classification KD benefits less
compared to LD.
Then, we impose the distillation on a larger range, \ie, VLR.
We can see that ``VLR LD" (the 5-th row of Table \ref{tab:separateKD}) can further
improve AP by +0.7 based on ``Main LD" (the 3-rd row).
However, we observe that further involving ``VLR KD" yields limited improvement (the 2-nd row and the 5-th row of Table \ref{tab:separateKD}) or even no improvement
(the last two rows of Table \ref{tab:separateKD}).
This again shows that localization knowledge distillation is more important and 
efficient than semantic knowledge distillation and
our divide-and-conquer distillation scheme, 
\ie, ``Main KD" + ``Main LD" + ``VLR LD", is complementary to VLR.

\begin{table}[!t]
	
	\centering
	
	\setlength{\abovecaptionskip}{3pt}
	\renewcommand\arraystretch{1}
	\caption{\textbf{Logit Mimicking {\em vs}. Feature Imitation.} ``Ours" means we use the separate distillation region  manner, \ie, conducing KD and LD on the main distillation region, and conducing LD on the VLR.
	The teacher is ResNet-101 and the student is ResNet-50 \cite{ResNet}. The results are reported on MS COCO val2017.}
	\begin{spacing}{1.1}
		\resizebox{0.47\textwidth}{!}{
			\begin{tabular}{l|c c c|c c c}
				\hline
				
				\hline
				Method & AP & AP$_{50}$ & AP$_{75}$ & AP$_{S}$ & AP$_{M}$ & AP$_{L}$  \\
                \hline
                
				Baseline (GFocal \cite{gfocal}) & 40.1 & 58.2 & 43.1 & 23.3 & 44.4 & 52.5 \\
				\hline
                FitNets \cite{FitNets} & 40.7 & 58.6 & 44.0 & 23.7 & 44.4 & 53.2 \\
                Inside GT Box  & 40.7 & 58.6 & 44.2 & 23.1 & 44.5 & 53.5 \\
                Main Region  & 41.1 & 58.7 & 44.4 & 24.1 & 44.6 & 53.6 \\
                Fine-Grained \cite{wang2019distilling} & 41.1 & 58.8 & 44.8 & 23.3 & 45.4 & 53.1 \\
                DeFeat \cite{defeat} & 40.8 & 58.6 & 44.2 & 24.3 & 44.6 & 53.7 \\
                GI Imitation \cite{GIbox} & 41.5 & 59.6 & 45.2 & 24.3 & 45.7 & 53.6 \\
                \hline
                Ours  & 42.1 & 60.3 & 45.6 & 24.5 & 46.2 & 54.8 \\
                Ours + FitNets  & 42.1 & 59.9 & 45.7 & \highlight{25.0} & 46.3 & 54.4 \\
                Ours + Inside GT Box  & 42.2 & 60.0 & 45.9 & 24.3 & 46.3 & 55.0 \\
                Ours + Main Region  & 42.1  & 60.0 & 45.7 & 24.6 & 46.3 & 54.7 \\
                Ours + Fine-Grained  & \highlight{42.4} & \highlight{60.3} & 45.9 & 24.7 & 46.5 & \highlight{55.4} \\
                Ours + DeFeat  & 42.2 & 60.0 & 45.8 & 24.7 & 46.1 & 54.4 \\
                Ours + GI Imitation  & \highlight{42.4} & \highlight{60.3} & \highlight{46.2} & \highlight{25.0} & \highlight{46.6} & 54.5 \\
				\hline
				
				\hline
			\end{tabular}
		}
	\end{spacing}
	\label{tab:compare}
\end{table}
\begin{figure}[t]
	\centering
	\setlength{\abovecaptionskip}{5pt}
	\includegraphics[width=0.47\textwidth]{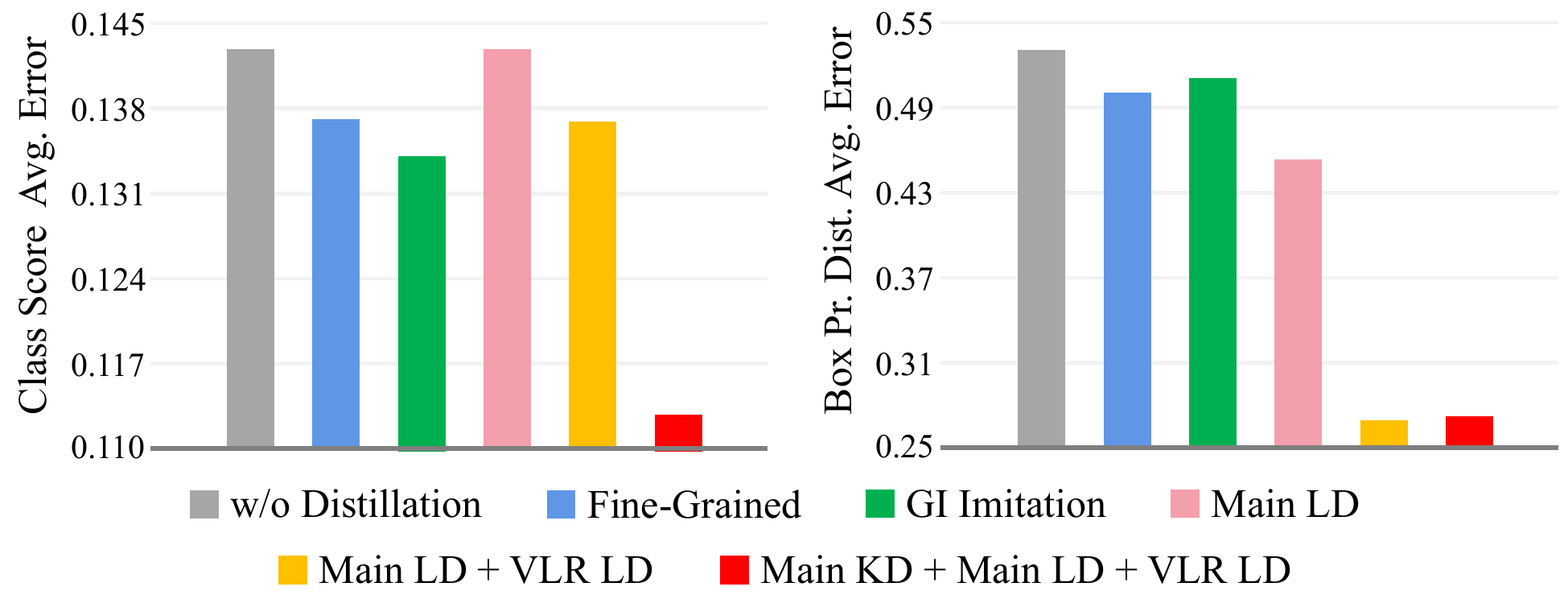}
	\caption{Visual comparisons of SOTA feature imitation and our LD. We show the average L1 error of classification scores and box probability distributions between teacher and student at the P4, P5, P6 and P7 FPN levels. The teacher is ResNet-101 and the student is ResNet-50. The results are evaluated on MS COCO val2017.}
	\label{fig:error}
\end{figure} 

\begin{figure*}[!t]
	\centering
	\setlength{\abovecaptionskip}{5pt}
	\includegraphics[width=0.98\textwidth]{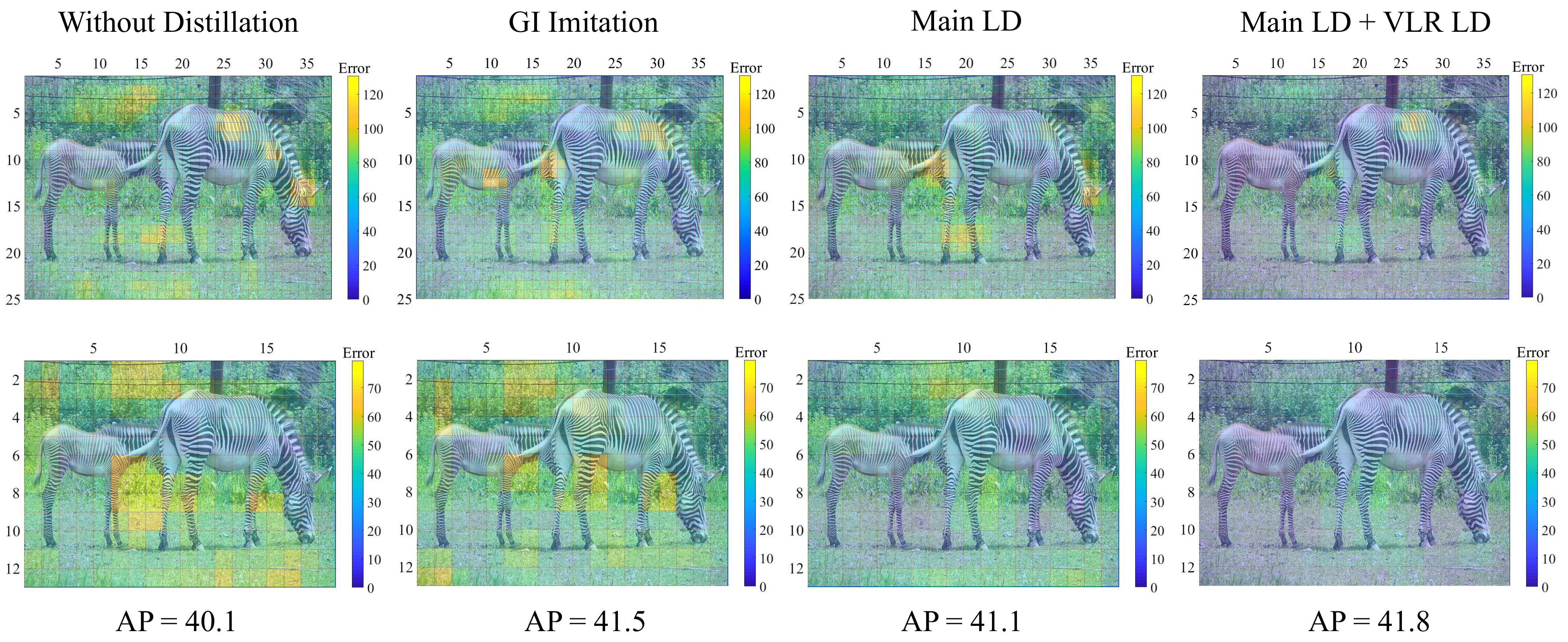}
	\vspace{-5pt}
	\caption{Visual comparisons between the state-of-the-art feature imitation and our LD. We show the per-location L1 error summation of the localization head logits between the teacher and the student at the P5 (first row) and P6 (second row) FPN levels. The teacher is ResNet-101 and the student is ResNet-50. We can see that compared to the GI imitation \cite{GIbox}, our method (Main LD + VLR LD) can
	significantly reduce the errors for almost all the locations.
	Darker is better. Best viewed in color.}
    \vspace{-5pt}
	\label{fig:visual}
\end{figure*}
\myPara{Logit Mimicking {\em vs}. Feature Imitation.}
We compare our proposed LD with several state-of-the-art feature imitation methods.
We adopt the separate distillation region manner, \ie, 
performing KD and LD on the main distillation region, and performing LD on the VLR.
Since modern detectors are usually equipped with FPN~\cite{FPN}, following previous works \cite{wang2019distilling,GIbox,defeat}, we re-implement their methods and impose all the feature imitations on multi-level FPN for a fair comparison.
Here, ``FitNets" \cite{FitNets} distills 
the whole feature maps.
``DeFeat" \cite{defeat} means the loss weights of feature imitation outside the GT boxes are larger than those inside GT boxes.
``Fine-Grained" \cite{wang2019distilling} distills the deep features on the close anchor box locations.
``GI Imitation" \cite{GIbox} selects the distillation regions according to the discriminative predictions of the student and the teacher.
``Inside GT Box" means we use the GT boxes 
with the same stride on the FPN layers as the feature imitation regions.
``Main Region" means we imitate the features within the main distillation region.

From Table \ref{tab:compare}, we can see that distillation
within the whole feature maps attains +0.6 AP gains.
By setting a larger loss weight for the locations outside the GT boxes (DeFeat \cite{defeat}), the performance is slightly better than that
using the same loss weight for all locations.
Fine-Grained \cite{wang2019distilling} focusing on the locations near GT boxes, produces 41.1 AP, which is comparable to the results of feature imitation using the Main Region.
GI imitation \cite{GIbox} searches the discriminative patches for feature imitation and gains 41.5 AP.
Due to the large gap in predictions between student and teacher, the imitation regions may appear anywhere.

Despite the notable improvements of these feature imitation methods, they do not explicitly consider the knowledge distribution patterns.
On the contrary, our method can transfer the knowledge via a separate distillation region manner, which directly produces 42.1 AP.
It is worth noting that our method operates on logits instead of features, indicating that logit mimicking is not inferior to feature imitation as long as adopting a proper distillation strategy, like our LD.
Moreover, our method is orthogonal to the aforementioned feature imitation methods.
Table~\ref{tab:compare} shows that with these feature imitation methods, our performance can be further improved.
Particularly, with GI imitation, we improve the strong GFocal baseline by +2.3 AP and +3.1 AP$_{75}$.

We further conduct an experiment to check the average error of classification score and box probability distribution, as shown in Fig. \ref{fig:error}.
One can see that the Fine-Grained feature imitation \cite{wang2019distilling} and GI imitation \cite{GIbox} reduce the two errors as expected, since the semantic knowledge and localization knowledge are mixed on feature maps.
Our ``Main LD" and ``Main LD + VLR LD" have comparable or larger classification score average errors than Fine-Grained \cite{wang2019distilling} and GI imitation \cite{GIbox} but lower box probability distribution average errors.
This indicates that these two settings
with only LD can significantly reduce the box probability distribution distance between the teacher and the student, while it is reasonable that they cannot reduce this error for classification head.
If we impose the classification 
KD on the main distillation region, 
obtaining ``Main KD + Main LD + VLR LD", 
both classification score average error and box probability distribution average error can be reduced.

We also visualize the L1 error summation of the localization head logits between the student and the teacher for each location at the P5 and P6 FPN levels.
In Fig. \ref{fig:visual}, comparing to ``Without Distillation", we can see that the GI imitation \cite{GIbox} does decrease the localization discrepancy between the teacher and the student.
Notice that we particularly choose a model (Main LD + VLR LD) with slightly better AP performance than GI imitation for visualization.
Our method can reduce this error more observably, and alleviate the localization ambiguity.

\begin{table}[!tb]
	\centering
	\renewcommand\arraystretch{1.05}
	\caption{Quantitative results of LD for lightweight detectors. The teacher is ResNet-101. The results are reported on MS COCO val2017.}	
	\vspace{-5pt}
	\begin{spacing}{1.1}
		\resizebox{0.47\textwidth}{!}{
			\begin{tabular}{c|c|c c c|c c c}
				\hline

				\hline
				Student & LD & AP & AP$_{50}$ & AP$_{75}$ & AP$_{S}$ & AP$_{M}$ & AP$_{L}$  \\
				\hline
				\multirow{2}{*}{ResNet-18}  &  & 35.8  & 53.1 & 38.2 &  18.9 &  38.9 &  47.9\\
				& \checkmark & 37.5  & 54.7 & 40.4 & 20.2 & 41.2 & 49.4 \\
                \hline
				\multirow{2}{*}{ResNet-34}&  & 38.9 & 56.6 & 42.2 &  21.5 &  42.8 & 51.4 \\
				&  \checkmark &  41.0 & 58.6 & 44.6 & 23.2 & 45.0 & 54.2 \\
				\hline
				\multirow{2}{*}{ResNet-50}&   & 40.1  & 58.2 & 43.1 &  23.3 & 44.4  & 52.5 \\
				& \checkmark  & 42.1  & 60.3 & 45.6 & 24.5 & 46.2  & 54.8 \\
				\hline

				\hline
			\end{tabular}
		}
	\end{spacing}
		\vspace{-0.1in}
		\label{tab:lightweight}
\end{table}

\myPara{LD for Lightweight Detectors.}
Next, we validate our LD with the separate distillation region manner, \ie, Main KD + Main LD + VLR LD, for lightweight detectors.
We select ResNet-101 with 44.7 AP provided by the mmDetection \cite{mmdetection} as our teacher to distill 
a series of lightweight students.
As shown in Table \ref{tab:lightweight}, our LD can stably improve the students ResNet-18, ResNet-34, ResNet-50 by +1.7, +2.1, +2.0 in AP, and +2.2, +2.4, +2.4 in AP$_{75}$, respectively.
From these results, we can conclude that our LD can stably 
improve the localization accuracy for all the students.

\begin{table}[!tb]
	\centering
	\renewcommand\arraystretch{1.05}
	\caption{Quantitative results of LD on various popular dense object detectors. The teacher is ResNet-101 and the student is ResNet-50. The results are reported on MS COCO val2017.}		
	\vspace{-5pt}
	\begin{spacing}{1.1}
		\resizebox{0.47\textwidth}{!}{
			\begin{tabular}{c|c|c c c|c c c}
				\hline

				\hline
				Student & LD & AP & AP$_{50}$ & AP$_{75}$ & AP$_{S}$ & AP$_{M}$ & AP$_{L}$  \\
				\hline
				\multirow{2}{*}{RetinaNet \cite{lin2017focal}}  &  & 36.9  & 54.3 & 39.8 &  21.2 &  40.8 &  48.4\\
				& \checkmark & 39.0  & 56.4 & 42.4 &  23.1 &  43.2 &  51.1\\
                \hline
				\multirow{2}{*}{FCOS \cite{FCOS}}&  & 38.6 & 57.2 & 41.5 &  22.4 &  42.2 & 49.8 \\
				&  \checkmark &  40.6 & 58.4 & 44.1 &  24.3 &  44.1 & 52.3 \\
				\hline
				\multirow{2}{*}{ATSS \cite{ATSS}}&   & 39.2  & 57.3 & 42.4 &  22.7 & 43.1  & 51.5 \\
				& \checkmark  & 41.6  & 59.3 & 45.3 & 25.2 & 45.2  & 53.3 \\
				\hline

				\hline
			\end{tabular}
		}
	\end{spacing}
		\vspace{-0.1in}
		\label{tab:other}
\end{table}

\myPara{Extension to Other Dense Object Detectors.}
Our LD is flexible to incorporate into other dense object detectors, either anchor-based or anchor-free type.
We employ LD with the separate distillation region manner to several recently popular detectors such as RetinaNet \cite{lin2017focal} (anchor-based), FCOS \cite{FCOS} (anchor-free) and ATSS \cite{ATSS} (anchor-based).
According to the results in Table \ref{tab:other}, LD can consistently improve $\thicksim$2 AP for these dense detectors.

\subsection{Comparison with the State-of-the-Arts}

We compare our LD with the state-of-the-art dense object detectors by using our LD to further boost GFocalV2 \cite{gfocalv2}.
For COCO val2017, since most previous works use ResNet-50-FPN backbone with single-scale $1\times$ training schedule (12 epochs) for validation, we also report the results under this setting for a fair comparison.
For COCO test-dev 2019, following previous work \cite{gfocalv2}, the LD models with $1333\times[480:960]$ multi-scale $2\times$ training schedule (24 epochs) are included.
The training is carried on a machine node with 8 GPUs using a batch size of
2 per GPU and initial learning rate 0.01 for a fair comparison.
During inference, single-scale testing ([$1333\times 800$] resolution) is adopted.
For different students ResNet-50, ResNet-101 and ResNeXt-101-32x4d-DCN \cite{xie2017aggregated,DCNv2}, we also choose different networks ResNet-101, ResNet-101-DCN and Res2Net-101-DCN \cite{gao2019res2net} as their teachers, respectively.

\begin{table}[!ht]
\caption{Compare with state-of-the-art methods on COCO \textit{val2017} and \textit{test-dev2019} . \textbf{TS}: Traning Schedule. '1$\times$': single-scale training 12 epochs. '2$\times$': multi-scale training 24 epochs.}
\label{tab:SOTA}
\vspace{-5pt}
\renewcommand{\arraystretch}{1.13}
\centering
\begin{adjustbox}{max width=\columnwidth}
\small{
\begin{tabular}{l|c|ccc|ccc}
    \hline
    
    \hline
    \textbf{Method} & \textbf{TS} & AP & AP$_{50}$ & AP$_{75}$ & AP$_{\it S}$ & AP$_{\it M}$ & AP$_{\it L}$ \\ 
    \hline
    
    \hline
    \multicolumn{8}{c}{\textbf{ResNet-50 backbone on val2017}} \\\hline
    RetinaNet \cite{lin2017focal}  & 1$\times$ & 36.9  & 54.3 & 39.8 &  21.2 &  40.8 &  48.4\\
    FCOS\cite{FCOS}		& 1$\times$	& 38.6 & 57.2 & 41.5 &  22.4 &  42.2 & 49.8 \\
    SAPD \cite{zhu2019soft} & 1$\times$ & 38.8 & 58.7  & 41.3  & 22.5  & 42.6 & 50.8 \\
    ATSS \cite{ATSS} & 1$\times$  & 39.2  & 57.3 & 42.4 &  22.7 & 43.1  & 51.5 \\ 
    BorderDet \cite{qiu2020borderdet} & 1$\times$  & 41.4 & 59.4 & 44.5 & 23.6 & 45.1 & 54.6 \\
    AutoAssign \cite{zhu2020autoassign} & 1$\times$ & 40.5 & 59.8 & 43.9 & 23.1 & 44.7 & 52.9 \\
    PAA \cite{kim2020probabilistic} & 1$\times$ & 40.4 & 58.4 & 43.9 & 22.9 & 44.3 & 54.0 \\
    OTA \cite{OTA} & 1$\times$ & 40.7 & 58.4 & 44.3 & 23.2 & 45.0 & 53.6 \\
    GFocal \cite{gfocal} & 1$\times$ & 40.1 & 58.2 & 43.1 & 23.3 & 44.4 & 52.5 \\
    GFocalV2 \cite{gfocalv2} & 1$\times$ & 41.1 & 58.8 & 44.9 & 23.5 & 44.9 & 53.3 \\  
    LD (ours) & 1$\times$ & \highlight{42.7} & \highlight{60.2} & \highlight{46.7} & \highlight{25.0} & \highlight{46.4} & \highlight{55.1} \\  
    \hline
    \multicolumn{8}{c}{\textbf{ResNet-101 backbone on test-dev 2019}} \\\hline
    RetinaNet \cite{lin2017focal}  & 2$\times$ & 39.1 & 59.1 & 42.3 & 21.8 & 42.7 & 50.2\\
    FCOS\cite{FCOS}		& 2$\times$	& 41.5 & 60.7 & 45.0 & 24.4 & 44.8 & 51.6 \\
    SAPD \cite{zhu2019soft} & 2$\times$ & 43.5 & 63.6 & 46.5 & 24.9 & 46.8 & 54.6 \\
    ATSS \cite{ATSS} & 2$\times$  & 43.6 & 62.1 & 47.4 & 26.1 & 47.0 & 53.6 \\ 
    BorderDet \cite{qiu2020borderdet} & 2$\times$  & 45.4 & 64.1 & 48.8 & 26.7 & 48.3 & 56.5 \\
    AutoAssign \cite{zhu2020autoassign} & 2$\times$ & 44.5 & 64.3 & 48.4 & 25.9 & 47.4 & 55.0 \\
    PAA \cite{kim2020probabilistic} & 2$\times$ & 44.8 & 63.3 & 48.7 & 26.5 & 48.8 & 56.3 \\
    OTA \cite{OTA} & 2$\times$ & 45.3 & 63.5 & 49.3 & 26.9 & 48.8 & 56.1\\
    GFocal \cite{gfocal} & 2$\times$ & 45.0 & 63.7 & 48.9 & 27.2 & 48.8 & 54.5\\
    GFocalV2 \cite{gfocalv2} & 2$\times$ & 46.0 & 64.1 & 50.2 & 27.6 & 49.6 & 56.5 \\  
    LD (ours) & 2$\times$ & \highlight{47.1} & \highlight{65.0} & \highlight{51.4} & \highlight{28.3} & \highlight{50.9} & \highlight{58.5} \\  
    \hline
    \multicolumn{8}{c}{\textbf{ResNeXt-101-32x4d-DCN backbone on test-dev 2019}} \\\hline
    SAPD \cite{zhu2019soft} & 2$\times$ & 46.6 & 66.6 & 50.0 & 27.3 & 49.7 & 60.7\\
    GFocal \cite{gfocal} & 2$\times$ & 48.2 & 67.4 & 52.6 & 29.2 & 51.7 & 60.2\\
    GFocalV2 \cite{gfocalv2} & 2$\times$ & 49.0 & 67.6 & 53.4 & 29.8 & 52.3 & 61.8  \\  
    LD (ours) & 2$\times$ & \highlight{50.5} & \highlight{69.0} & \highlight{55.3} & \highlight{30.9} & \highlight{54.4} & \highlight{63.4} \\  
    \hline
    
    \hline
\end{tabular}}
\end{adjustbox}
\end{table}

As shown in Table \ref{tab:SOTA}, our LD improves the AP score of the SOTA GFocalV2
by +1.6 and the AP$_{75}$ score by +1.8 when using the ResNet-50-FPN backbone.
When using the ResNet-101-FPN and ResNeXt-101-32x4d-DCN with multi-scale $2\times$ training, we achieve the highest AP scores, 47.1 and 50.5 , which outperform all existing dense object detectors under the same backbone, neck and test settings.
More importantly, our LD does not introduce any additional network parameters or computational overhead and thus can guarantee exactly the same inference speed as GFocalV2.

\section{Conclusion}\label{sec:conclusion}
In this paper, we propose a flexible localization distillation for dense object detection and design a valuable localization region to distill the student detector in a separate distillation region manner.
We show that 1) logit mimicking can be better than feature imitation;
and 2) the separate distillation region manner for transferring the classification and localization knowledge is important when distilling object detectors.
We hope our method could provide new research intuitions for the object detection community to develop better distillation strategies.
In addition, the applications of LD to sparse object detectors (DETR \cite{DETR} series) and other relevant fields, \eg, instance segmentation, object tracking and 3D object detection, warrant future research.

\myPara{Acknowledgment}
This work is funded by the 
National Key Research and Development Program of China (NO. 2018AAA0100400) and
NSFC (NO. 61922046, NO. 62172127).

{\small
	\bibliographystyle{ieee_fullname}
	\bibliography{egbib}
}

\appendix

\renewcommand{\thesection}{A\arabic{section}}
\renewcommand{\thetable}{A\arabic{table}}
\renewcommand{\thefigure}{A\arabic{figure}}

\setcounter{table}{0}
\section{Implementation Details}

The training is carried on 8 GPUs with batch size 2 images per GPU.
The total training epochs are set to 12 for 1$\times$ training schedule 
and 24 for 2$\times$ training schedule following most previous work.
The initial learning rate is 0.01 and a linear warm-up strategy is used 
for the first 500 iterations.
The learning rate decreases by a factor of 10 after the 8-th and 11-th epoch for 1$\times$ training schedule, while after the 16-th and 22-th epoch 
for 2$\times$ training schedule.
For ablation studies, we adopt single-scale training and 
test with $1333\times800$ resolution.

We also provide experiment results on another popular object detection benchmark, \ie, PASCAL VOC \cite{voc}.
We use VOC 07+12 training protocol, \ie, the union of VOC 2007 trainval set and VOC 2012 trainval set (16551 images) for training, and VOC 2007 test set (4952 images) for evaluation.
The initial learning rate is 0.01 and the total training epochs are set to 4.
The learning rate decreases by a factor of 10 after the 3-rd epoch.
We report results in terms of AP and 5 mAPs under different IoU thresholds.

\section{More Method Studies}

\myPara{Individual LD.}
To investigate the performance of LD, we do not use the ground-truth bounding boxes
for training the student detector by disabling the bounding box regression loss $\mathcal{L}_\text{reg}$ and the distribution focal loss $\mathcal{L}_\text{DFL}$
in Eqn.~(5).
From Table \ref{tab:only-LD}, we can see that when we only use LD 
within the main distillation region to train the student ResNet-18, we attain
36.4 AP and 39.3 AP$_{75}$.
These results are surprisingly higher than those using
$\mathcal{L}_\text{reg}$ and $\mathcal{L}_\text{DFL}$ (the first row), 
reflecting the capability of LD in helping the student to learn rich localization knowledge.
With both $\mathcal{L}_\text{reg}$ and $\mathcal{L}_\text{DFL}$ added, 
only 0.1 AP is increased, indicating that the probability distribution learned 
by the teacher detector is a better localization supervision than the 
ground-truth bounding box annotations.
\begin{table}[!t]\small
	\centering
	\setlength{\tabcolsep}{2pt}
	\setlength{\abovecaptionskip}{3pt}
	\caption{Evaluation of \textbf{Individual LD} on the main distillation region. The results are reported on MS COCO val2017. \textbf{R}: ResNet \cite{ResNet}.}
	\begin{spacing}{1.1}
		\resizebox{0.48\textwidth}{!}{
		\begin{tabular}{ccc|ccc|ccc|ccc|ccccc}
			\hline
			
			\hline
			& Teacher &&& Student   &&& $\mathcal{L}_\text{LD}$ &&&$\mathcal{L}_\text{reg}$ and $\mathcal{L}_\text{DFL}$ &&& AP  &&& AP$_{75}$ \\
			\hline
			& \multirow{3}{*}{\textbf{R}-101}&&& \multirow{3}{*}{\textbf{R}-18} &&& &&& \checkmark &&& 35.8  &&& 38.2 \\
			& &&& &&&   \checkmark &&&  &&& 36.4  &&& 39.3\\
			& &&& &&&   \checkmark &&& \checkmark &&& 36.5  &&& 39.3\\
			\hline
			
			\hline
		\end{tabular}}
		\label{tab:only-LD}
	\end{spacing}
\end{table}

\myPara{Self-LD.}
In KD, the student model $\bm S$ is generally lighter than the teacher model $\bm T$
so as to learn compact and efficient models.
Recently, self-distillation has been consistently observed to have a positive effect 
in classification \cite{furlanello2018born,zhang2019your}.
For object detection, it is also positive to see that Self-LD with $\bm S = \bm T$ can also bring performance gains.
For the reason why self-distillation can facilitate the model accuracy, \cite{mobahi2020self} firstly unveils the mystery by theoretically analyzing self-distillation in Hilbert Space.
It has been proven that a few rounds of self-distillation can reduce over-fitting 
since it induces the regularization.
However, continuing self-distillation may lead to under-fitting, 
and we simply perform self-LD once.
As listed in Table \ref{tab:self-LD}, Self-LD on the main distillation region consistently boosts the performance by +0.3 AP, +0.3 AP$_{75}$ for ResNet-18, +0.5 AP, +0.7 AP$_{75}$ for ResNet-50, and +0.6 AP, +0.7 AP$_{75}$ for ResNeXt-101-32x4d-DCN.
Self-LD shows the universality of our LD that the localization knowledge can still be transferred when the teacher model is with the same scale as the student model.
\begin{table}[!t]
    \small
	\centering
\setlength{\abovecaptionskip}{3pt}
\renewcommand\arraystretch{1}
	\caption{Quantitative results of \textbf{Self-LD} on the main distillation region. The results are reported on MS COCO val2017.}
	\begin{spacing}{1.1}
		\resizebox{0.48\textwidth}{!}{
		\begin{tabular}{c|c|ll}
		    \hline
		
		    \hline
			 Detector   & Self-LD  & AP   & AP$_{75}$ \\
			\hline
			 \multirow{2}{*}{ResNet-18} &   & 35.8   & 38.2 \\
			 &   \checkmark   & \textbf{36.1 (+0.3)}  & \textbf{38.5}\\
			\hline
			 \multirow{2}{*}{ResNet-50}  &    & 40.1  & 43.1\\
			&   \checkmark   & \textbf{40.6 (+0.5)}  & \textbf{43.8}\\
			\hline
			 \multirow{2}{*}{ResNeXt-101-32x4d-DCN} &     & 46.9   & 51.1 \\
			&    \checkmark   & \textbf{47.5 (+0.6)}  & \textbf{51.8}\\
		    \hline
		
		    \hline
		\end{tabular}
        }
		\label{tab:self-LD}
	\end{spacing}
\end{table}

\begin{table}[htb!]\small
	\centering
	\setlength{\tabcolsep}{4pt}
	\setlength{\abovecaptionskip}{2pt}
	\renewcommand\arraystretch{1.1}
	\caption{Effect of using our LD on model size, FLOPs and FPS. FPS is measured using a RTX 2080 Ti GPU and averaged over 3 runs.}
	\resizebox{0.47\textwidth}{!}{
		\begin{tabular}{c c c c cc c c}
			\hline
			
			\hline
			  & LD & image size & \#param. & FLOPs & FPS & AP  \\
            \hline
			 \multirow{2}{*}{RetinaNet} & & $1333\times800$ & 37.74M & 239.32G & 21.0 & 36.9 \\
			  & \checkmark & $1333\times800$ & 39.07M & 267.64G & 19.6 & 39.0 \\
            \hline
			 \multirow{2}{*}{FCOS} & & $1333\times800$ & 32.02M & 200.50G & 22.3 & 38.6 \\
			  & \checkmark & $1333\times800$ & 32.17M & 203.60G & 22.4 & 40.6 \\
			  \hline
			  \multirow{2}{*}{ATSS} & & $1333\times800$ & 32.07M & 205.21G & 21.9 & 39.2 \\
			  & \checkmark & $1333\times800$ & 32.22M & 208.36G & 21.9 & 41.6 \\
			  \hline
			  \multirow{2}{*}{GFocal} & & $1333\times800$ & 32.22M & 208.31G & 21.9 & 40.1 \\
			  & \checkmark & $1333\times800$ & 32.22M & 208.31G & 21.9 & 42.1 \\

			\hline
			
			\hline
		\end{tabular}}
	\label{tab:infspeed}
\end{table}
    
\begin{table}[htb!]
	\small
	\centering
	\setlength{\tabcolsep}{4.5pt}
	\renewcommand\arraystretch{1.2}
	\setlength{\abovecaptionskip}{3pt}
	\caption{Quantitative results of LD on the main distillation region. The results are reported on the test set of PASCAL VOC 2007.}
	\begin{tabular}{c|c|cccc}
		\hline
		
		\hline
		Teacher &  Student   & AP & AP$_{50}$  & AP$_{70}$  & AP$_{90}$ \\
		\hline
		-- &  \multirow{5}{*}{ResNet-18}   & 51.8 & 75.8  & 62.9 &  20.6 \\
		\cline{1-1}\cline{3-6}
		ResNet-34 & & 52.9  & 75.7  & \textbf{64.5}  & 22.0 \\
		\cline{1-1}\cline{3-6}
		ResNet-50 & & 52.8 & 75.4  & 64.0 & \textbf{23.0} \\
		\cline{1-1}\cline{3-6}
		ResNet-101 &  & \textbf{53.0}  & \textbf{75.9}  & 64.0 &  22.7 \\
		\cline{1-1}\cline{3-6}
		ResNet-101-DCN & & 52.8 & 75.3 & 64.2  & 22.1 \\
		\hline

		\hline
	\end{tabular}
	\label{tab:voc}
\end{table}

\begin{figure*}[htb!]
	\centering
	\small
	\setlength{\tabcolsep}{1pt}
	\setlength{\abovecaptionskip}{3pt}
		\includegraphics[width=\textwidth]{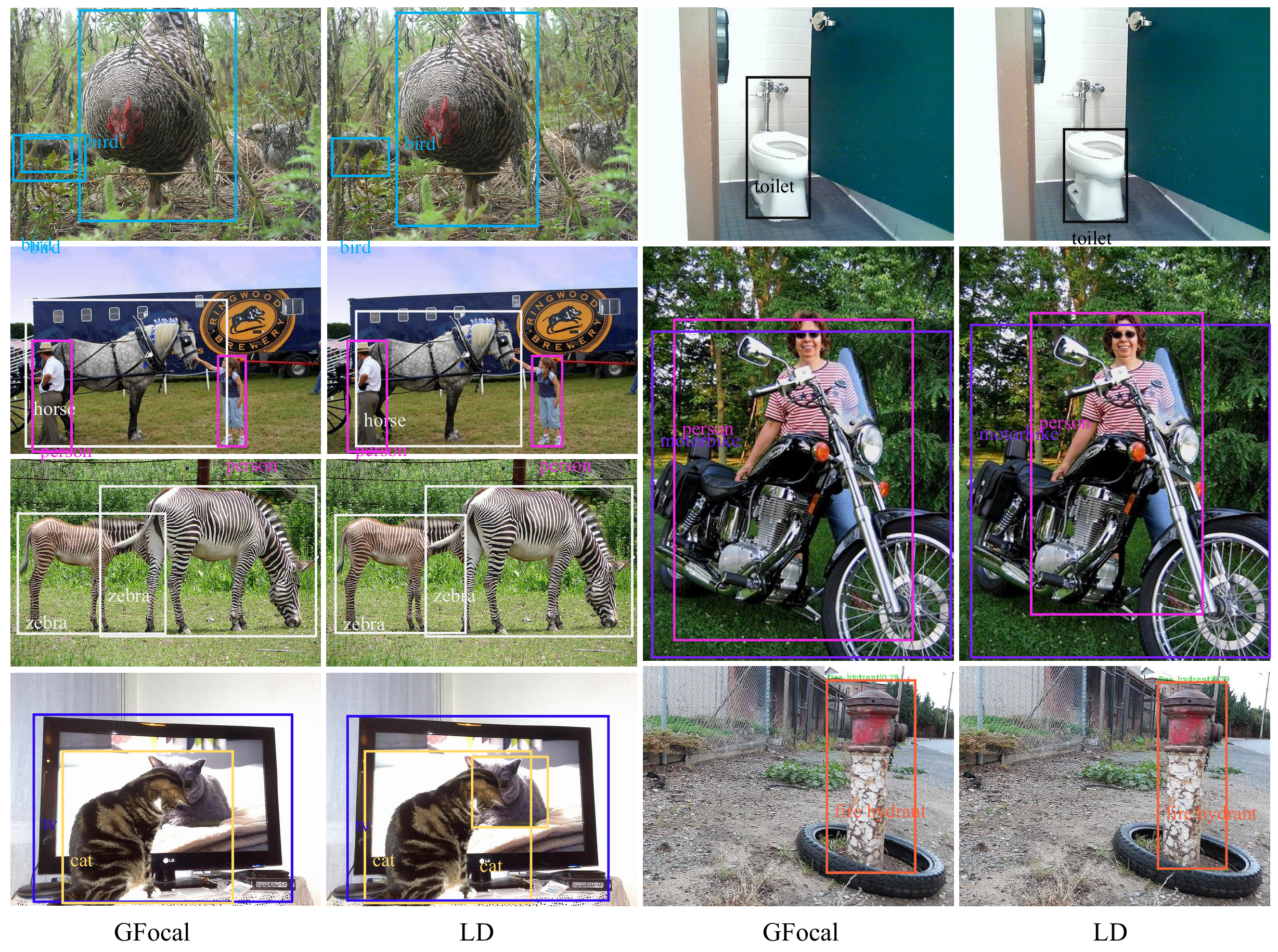}
	\caption{ Detection results by original GFocal and LD.
		Our LD makes detected boxes become more accurate. }
	\label{fig:main-detection}
\end{figure*}

\noindent\textbf{Inference Speed.}
Since our method needs to transform the bbox representation to probability distribution, the only modification to the detector lies in the localization head output channel, which is from $H\times W\times4$ to $H\times W\times4n$.
We also investigate this effect on model size, computations (FLOPs), and running speed (FPS) in Table \ref{tab:infspeed}.
We can see that our method can significantly improve the performance with negligible increase on model size and FLOPs for both FCOS and ATSS.
While for RetinaNet, it leads to a slight FPS drop.
This is mainly because RetinaNet uses 9 anchor boxes per location.
The modification of the localization head output channel of RetinaNet is from $H\times W\times(9\times4)$ to $H\times W\times(9\times4n)$, 8 times more than those of FCOS and ATSS.
As for GFocal, we obtain a free improvement.

\myPara{LD for Lightweight Detectors on PASCAL VOC.}
We further conduct experiments on PASCAL VOC to check the effectiveness of our LD.
Here, the main distillation region is adopted.
From Table \ref{tab:voc}, we can see that our LD consistently boost the performance for the student ResNet-18.
Notice that our LD performs significantly better than the baseline for the AP metrics under high IoU threshold, like AP$_{90}$.

\section{More Visualization}
\begin{figure*}[htb!]
	\centering
	\small
	\setlength{\tabcolsep}{1pt}
	\setlength{\abovecaptionskip}{3pt}
	\begin{tabular}{cccccccccc}
		\includegraphics[width=0.24\textwidth]{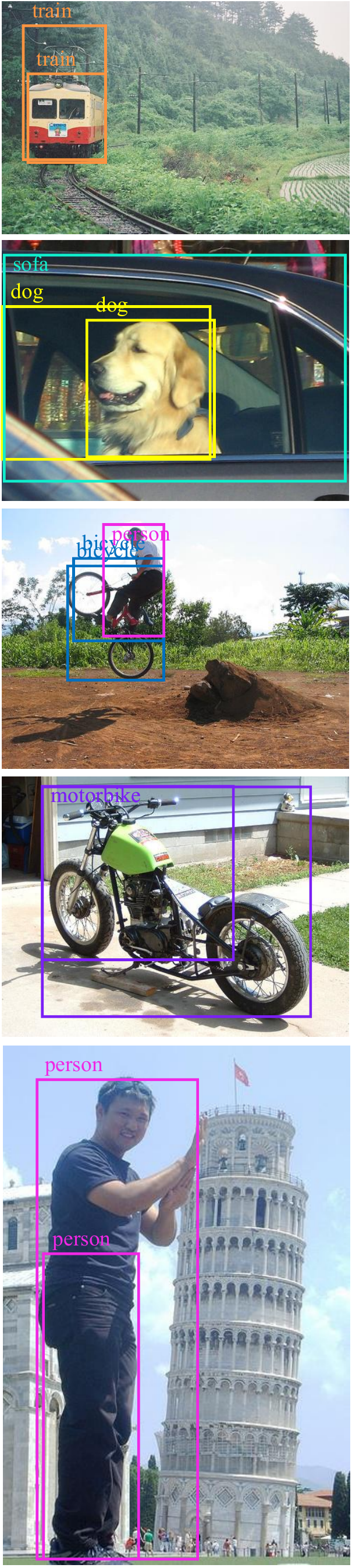}&
		\includegraphics[width=0.24\textwidth]{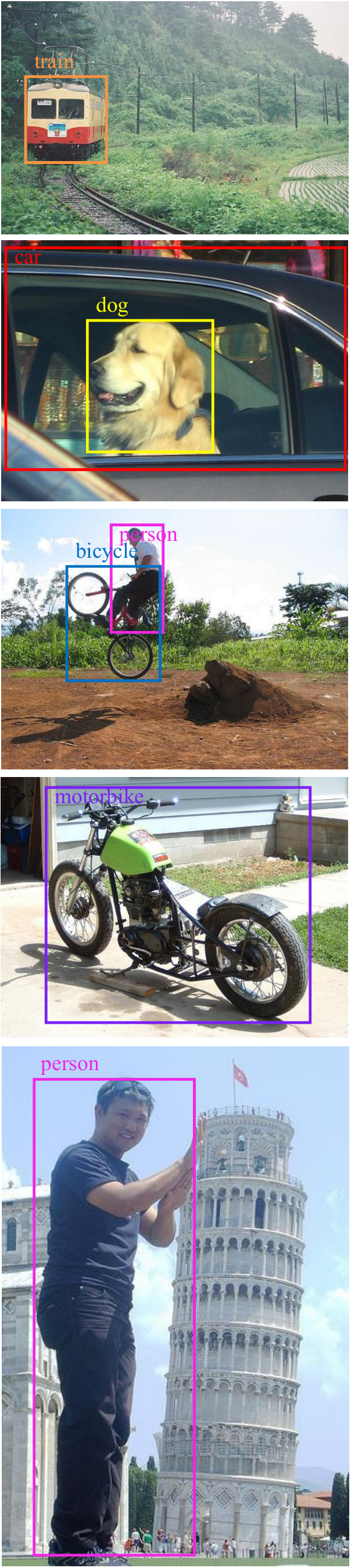}&
		\includegraphics[width=0.24\textwidth]{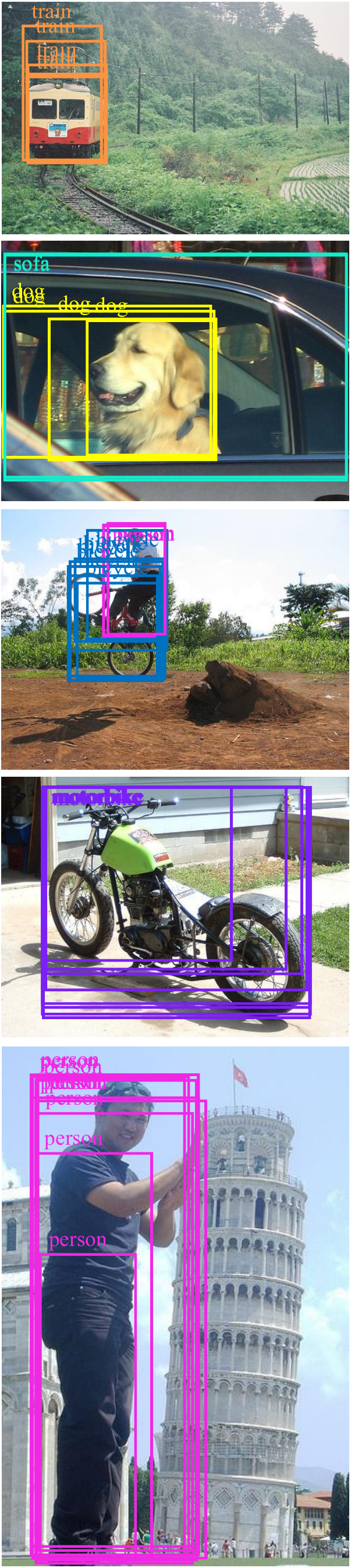}&
		\includegraphics[width=0.24\textwidth]{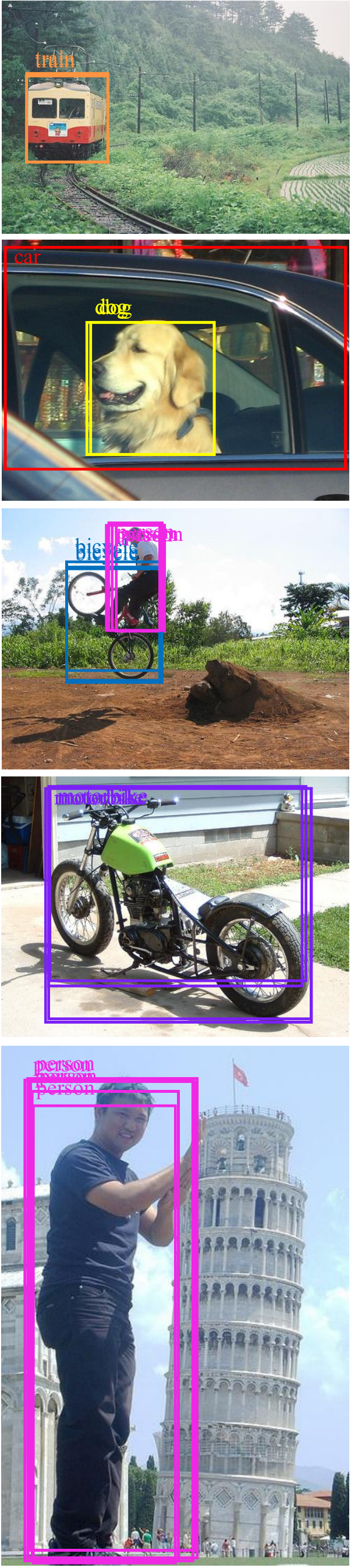}\\
		GFocal & LD & GFocal `0.95' & LD `0.95'\\
	\end{tabular}
	\caption{Detection results by GFocal and LD.
		`0.95' means that default NMS threshold 0.6 in GFocal is increased to 0.95.
		Detected boxes by original GFocal are with varying localization qualities, and redundant boxes may still survive after default NMS.
		In contrast, our LD contributes to improve localization quality of detected boxes, among which redundant ones are easy to be suppressed by default NMS.
	}
	\label{fig:detection}
\end{figure*}

First, we provide more detection results by GFocal and our LD in Fig. \ref{fig:main-detection}, from which one can see more accurate localization boxes are obtained by LD. 
Then, we present more detection results by GFocal and LD with different thresholds of NMS, as shown in Fig. \ref{fig:detection}.
Since our LD contributes to improve localization quality of detected boxes (referring to the results of \emph{LD `0.95' v.s. GFocal `0.95'}), redundant boxes can be suppressed by the default NMS (referring to the results of \emph{GFocal v.s. LD}). 

\end{document}